\documentclass[preprint,3p,times]{elsarticle}

\usepackage{amssymb}
\usepackage{amsthm}
\usepackage{gensymb}
\usepackage{lineno}
\usepackage{caption}
\usepackage[fleqn]{amsmath}
\usepackage{multirow}

\usepackage{hyperref}
\biboptions{sort&compress}
\hypersetup{colorlinks = true, allcolors = blue, pdfauthor=author}
\usepackage[nameinlink,capitalise]{cleveref}

\modulolinenumbers[5]
\begin{document}

\begin{frontmatter}



\title{Elastic energy storage of spring-driven jumping robots}


\author[inst1]{John Lo \corref{cor1}}
\cortext[cor1]{Corresponding author.}
\ead{kwokcheungjohn.lo@manchester.ac.uk}

\author[inst1,inst2]{Ben Parslew}
\affiliation[inst1]{organization={Department of Fluid and Environment},
            addressline={The University of Manchester}, 
            city={Manchester},
            postcode={M13 9PL}, 
            country={United Kingdom}}

\affiliation[inst2]{organization={International School of Engineering, Faculty of Engineering
},
            addressline={Chulalongkorn University}, 
            city={Bangkok},
            postcode={10330}, 
            country={Thailand}}

\begin{abstract}
Spring-driven jumping robots use an energised spring for propulsion, while the onboard motor only serves as a spring-charging source. A common mechanism in designing these robots is the rhomboidal linkage, which has been combined with linear springs (spring-linkage) to create a nonlinear spring, thereby increasing elastic energy storage and jump height for a given motor force. The effectiveness of this spring-linkage has been examined for individual designs, but a generalised design theory of this class of system remains absent. This paper presents an energetics analysis of the spring-linkage and provides insight into designing an ideal constant force spring, which stores the maximum energy for a given motor force. A quasi-static analysis shows that the force-displacement relationship of the spring-linkage changes with the orientation and type of the spring, but is independent of the linkage scale. Combining different types and orientations of springs within the linkage enables higher elastic energy storage than using single springs. Placing two translational springs at the diagonals of the rhomboidal linkage creates an ideal spring that could increase the jump height of prior robots by 50-160\%.
\end{abstract}

\begin{keyword}
Jumping robot \sep Elastic energy storage \sep Rhomboidal linkage \sep Energy efficiency \sep Spring-linkage

\end{keyword}

\end{frontmatter}


\section{Introduction}
\label{sec:intro}

\begin{figure}[t]
\centering
\includegraphics[scale=0.97]{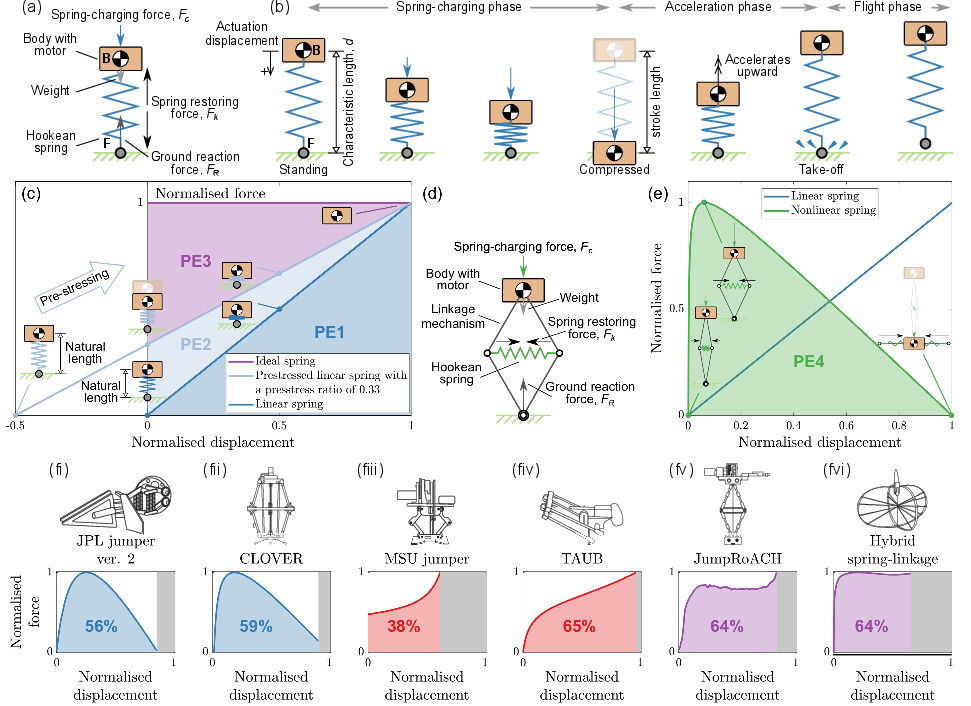}
\caption{An example of (a) a prismatic jumping robot driven by a linear spring (e.g.\citep{JPL_Fiorini_1999}]) and (b) the phases of the jumping process. (c) Force-displacement relationships of ideal, linear and pre-stressed linear springs; the pre-stress ratio is the ratio of the natural spring length of the pre-stressed spring to the system characteristic length. (d) Schematic of the spring-driven jumping robot proposed by \citep{JPL_Hale_2000}, which combines a linear spring with a rotational linkage to create a nonlinear spring; spring shown at its natural length. (e) Force-displacement relationships of a linear spring, and the nonlinear spring systems from (d). (f) Force-displacement relationships of some previous jumping robots with nonlinear spring-linkage mechanisms with different spring types and locations. (i)-(ii) use horizontal translational springs \citep{JPL_Hale_2000,CLOVER_2022}. (iii)-(iv) use rotational springs \cite{MSU_Zhao_2013,TAUB_Zaitsev_2015}. (v)-(vi) Combine rotational and translation springs in an attempt to create an idealised constant force spring \citep{Jumproach_Jung_2019,NatureJ_Hawkes_2022}. The greyed area is the unattainable region where the linkage could be further compressed but was prevented due to other physical constraints. The force is normalised by the peak spring-charging force in each example, and the displacement is the translational displacement of the body measured from the initial condition and is normalised by the characteristic length of each system.}
\label{fig:intro}
\end{figure}

The mechanical performance of motor-driven jumping robots is limited by the motor maximum power and velocity \citep{Ilton_2018,NatureJ_Hawkes_2022}. To tackle this issue, a mechanical spring can be used to increase the mechanical power available for jumping (\cref{fig:intro}a). In spring-driven jumpers the motor does work to increase the elastic potential energy stored in the spring, which is then released and converted into kinetic energy for jumping, as shown in \cref{fig:intro}b. By doing so, the jump height and power output of the spring-driven system are decoupled from the motor’s velocity and power outputs \citep{Ilton_2018,NatureJ_Hawkes_2022}.

Spring-driven jumping robots have been developed with various structures and arrangements of elastic elements \citep{JReview_Zhang_2017,JReview_Mo_2020,JReview_Ribak_2020,JReview_Zhang_2020}. A common goal for these systems is to achieve a high \textit{mechanical-elastic} energy conversion efficiency from the mechanical work done by a single motor stroke to the elastic energy stored in the spring, and then a high \textit{elastic-kinetic} energy conversion efficiency from the stored energy to the kinetic energy. This study will focus on mechanical-elastic energy conversion. 

Ideal mechanical-elastic efficiency is obtained theoretically using a constant force spring, where the applied force is independent of the displacement. The maximum elastic potential energy stored in the idealised system is the peak force multiplied by the spring length (the areas PE1, PE2 and PE3 in \cref{fig:intro}c). In comparison, a linear spring, such as that shown in \cref{fig:intro}a, can only store half of the maximum elastic potential energy for the same applied force and length as the ideal system. This paper will consider the physically realisable linkage spring arrangements and their energetics in comparison to idealised systems.

A conceptually simple approach to increasing the energy stored in a linear spring system for a given charging force is to pre-stress the spring. This involves using a linear spring with a natural length longer than the system characteristic length. The spring is pre-stressed by an external force before placing it into the jumping drivetrain. The pre-stressing phase shifts the force-displacement curve of the linear spring upward in \cref{fig:intro}c and subsequently increases the elastic potential energy storage for a given charging force. Pre-stressed systems are potentially a useful approach in the design of jumping systems. However, a caveat of this method is that to increase the elastic energy storage of the system for a given force, the natural spring length has to be increased. This implies that achieving the idealised mechanical-elastic energy efficiency of an ideal spring would require an infinitely long spring, which is practically unfeasible. Another practical limitation is the coils of the metallic helical springs would eventually come into contact under compression, which reduces the working length of the spring. The pre-stressed ratio of the system is constrained by the minimum working length of the physical spring.    As a more common approach, jumping robots adopt nonlinear springs, which also have a higher elastic energy storage than a linear spring for a given charging force. 

Nonlinear springs have been made previously by combining a planar closed-chained four-bar linkage mechanism with a linear spring \citep{JPL_Hale_2000}, as shown in \cref{fig:intro}d. The resulting spring-linkage system exhibits a nonlinear force-displacement relationship, as shown in \cref{fig:intro}e. This mechanical configuration was first proposed to replace the linear spring-driven jumping robot \citep{JPL_Fiorini_1999} that suffered energy lost from \textit{premature take-off} \citep{JPL_Hale_2000}, which is where the robot leaves the ground while a portion of elastic energy is still stored within the spring, and has not been converted to kinetic energy before take-off.  The nonlinear spring did not mitigate premature take-off \citep{NatureJ_Hawkes_2022,MSU_Zhao_2013,PEAnSEA_Hong_2020,LO_2021,Jollbot_Armour_2008} due to the rotational and unsprung masses inherent in the linkages \citep{LO_Dynamics_2023}. However, the spring-linkage did increase mechanical-elastic conversion efficiency to yield higher jumps than previous systems, despite the premature take-off \citep{LO_2021}. Subsequently, a whole class of similar nonlinear spring mechanisms were developed with various spring arrangements (\cref{fig:intro}fi-fvi), including rotational springs \citep{MSU_Zhao_2013,LO_2021,Jollbot_Armour_2008,TAUB_Zaitsev_2015,Springtail_Ma_2021,TSJ_Zhang_2018}, translational springs in different alignments \citep{JPL_Hale_2000,PEAnSEA_Hong_2020,JPL_Burdick_2003,FlapJ_Truong_2019,Jumproach_Jung_2016,FleaJ_Koh_2013,ScoutJ_Zhao_2009,ESJ_Bai_2018,Multimobat_Woodward_2014,CLOVER_2022,SurveillanceJ_Song_2009,SphericalJ_Chang_2022} and a combination of the two \citep{NatureJ_Hawkes_2022,Jumproach_Jung_2019}. The highest mechanical-elastic conversion efficiency shown previously is around 65\% (\cref{fig:intro}fiv), while some nonlinear springs only have a conversion efficiency of 38\% – lower than that of a linear spring system (\cref{fig:intro}fiii). The overall technological evolution of nonlinear springs for jumping systems has not been straightforward  and does not show a steady increase in mechanical-elastic conversion efficiency over time. For example, \citep{JPL_Hale_2000} developed in 2000 achieved an efficiency of 56\%, whereas \citep{FleaJ_Koh_2013} in 2013 attained only 40\%. The development of spring-driven jumping robots has not shown a steady progression but instead has focused on individual or bespoke designs for specific missions or applications \citep{JReview_Zhang_2017,JReview_Mo_2020,JReview_Ribak_2020,JReview_Zhang_2020}.  The contribution of this work is to take a methodical approach to characterise the mechanical-elastic conversion efficiency of nonlinear spring systems. This will be achieved by building a fundamental understanding of how conceptually simple spring-linkage systems can be arranged to achieve an ideal, or near-ideal mechanical-elastic energy conversion.

\section{Topology and spring-charging process of spring-linkage systems}
\label{sec:topo}
The prismatic linear spring system in \cref{fig:intro}a obeys Hooke’s law and has an elastic potential energy storage limited to 50\% of that of an ideal spring. This energy constraint is not imposed by the linear spring alone but is due to the topology of the combined spring and prismatic linkage. A linear spring with different linkages can store different amounts of elastic potential energy. 

\begin{figure}[t]
\centering
\includegraphics[scale=1]{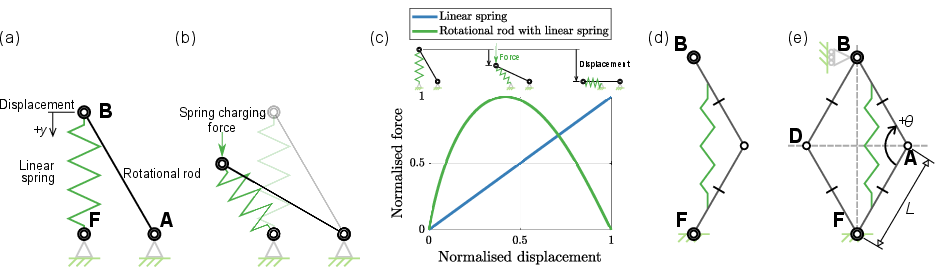}
\caption{(a) A rotational link with a linear translational spring being (a) at rest and (b) compressed by an external spring-charging force. (c) The spring charging force is plotted against the vertical displacement.  Note that (a)-(c) show the simplest topological arrangement of the translational spring with rotation linkage that exhibits a nonlinear force-displacement property. (d) An open-chained two-bar linkage is formed by adding an extra rotational link to (a). To avoid the multiple degrees of freedom, (d) can be mirrored into (e) a closed-chained four-bar linkage with a single translational degree of freedom, referred to as the \textit{rhomboidal linkage} in this study.}
\label{fig:topo}
\end{figure}

\cref{fig:topo}a is an example of the simplest spring linkage topology that exhibits a nonlinear force-displacement relationship. It is a nonlinear spring formed using a linear translational spring and a single link, with a single rotational degree of freedom. The link is connected to the ground with a revolute joint at A. B is connected to a translational linear spring via a revolute joint. Point F on the spring is connected to the ground through a revolute joint. Initially, this system is at equilibrium and the spring remains at its natural length, with the major axis perpendicular to the ground plane. As a force parallel to the $+y$ direction is applied at joint B the link will rotate and the linear spring will rotate and compress. Unlike the prismatic systems in \cref{fig:intro}a, the translational displacement of the spring in \cref{fig:topo} does not align with the y-axis and is not directly proportional to the applied spring-charging force vector parallel to the y-axis; the nonlinear force-length relationship in \cref{fig:topo}c  shows how the pairing of rotational link and linear spring forms an example nonlinear spring. This study will show how the force-length profile can be modified using linkages and springs, and how this influences elastic energy storage within the system.

While simple to comprehend, the example in \cref{fig:topo}a-c cannot be readily implemented in a jumping system as the link and spring are connected to the ground. The open-chain two-bar linkage in \cref{fig:topo}d forms a nonlinear spring without requiring a connection to the ground. While this linkage can be used as the basis for a jumping robot design \citep{TAUB_Zaitsev_2015}, the lack of symmetry results in lateral and angular acceleration during take-off.  A conceptually simpler model that will be applied here to understand the fundamentals of nonlinear springs uses a symmetrical closed-chain four-bar linkage (\cref{fig:topo}e), which only has a single degree of freedom. This linkage has also proven to be a practical solution to designing jumping robots \citep{JReview_Zhang_2017,JReview_Mo_2020,JReview_Ribak_2020,JReview_Zhang_2020}.

The closed-chained four-bar linkage mechanism is modelled as four bars of lengths $L$, connected by four revolute joints – A, B, D, and F, and is hereafter defined as rhomboidal linkage. The linkage model has only a single degree of freedom and is considered to have a single point contact with the ground at F; in practice, a finite base of support is required to avoid toppling and maintain the single translational degree of freedom, however this does not influence the mechanical-elastic conversion efficiency and so is neglected here. The characteristic length of the rhomboidal linkage is given as,
\begin{equation}
    d=2L,\label{eq_Topo_d}
\end{equation}
which is also the maximum displacement between joints B and F. Note that if the bars have different lengths, joint B cannot be displaced to joint F in order to maximise the mechanical work of the spring-charging force.

\begin{figure}[t]
\centering
\includegraphics[scale=1]{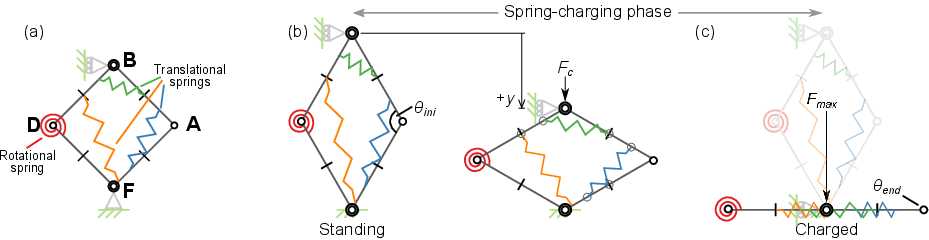}
\caption{A rhomboidal linkage with multiple translational and rotational springs shown at arbitrary attachments points to the linkage. The spring charging phase of (a) from $\theta_{ini}$=120\degree to $\theta_{end}=0\degree$. The key states are: (b) standing and (c) charged.}
\label{fig:spring_charging_phase}
\end{figure}

Joints B and F are referred to hereafter as the body and foot respectively.  Revolute joints A and D are referred to as the knee joints; the knee angle, $\theta$, is the angle between the two segments. If the spring mass is assumed negligible, the translational and rotational springs can be placed at any arbitrary position within the rhomboidal linkage, including the revolute joints, as shown in \cref{fig:spring_charging_phase}a. But note that in practice, the distribution of the spring mass within the linkage influences the jumping dynamics and energy efficiency of the system during the acceleration phase \citep{LO_Dynamics_2023}. 

To charge the spring-linkage, the spring charging process begins with the system in a standing posture, where the spring is at its natural length. The system is in contact with the ground via its foot (joint F); the standing knee angle is defined as the initial standing angle $\theta_{ini}$ (\cref{fig:spring_charging_phase}b). A charging force, $F_c$, is applied on the body (joint B) to compress the linkage system; the charging force is typically applied by an electric motor. The displacement of the body is expressed as a function of knee angle,
\begin{equation}
    y=d(\sin\frac{\theta_{ini}}{2}-\sin\frac{\theta}{2}).\label{eq_y}
\end{equation}

The work done by the charging force in displacing the spring is stored as elastic potential energy in the spring. The charging process will stop either when the spring restoring force reaches the peak charging force, $F_{max}$, or when the linkage mechanism has been fully depressed (where $\theta=0\degree$ and links BA and BD have rotated to become parallel with links FA and FD, respectively); in both cases the system is classified as being charged. When charged the knee angle is defined as $\theta_{end}$ (\cref{fig:spring_charging_phase}c). The stroke length is the displacement of joint B measured from full extension ($\theta=180\degree$) to fully depressed. During the process, inertial forces are neglected in the compression phase where the acceleration is considered to be negligible. Frictional forces are also neglected. Therefore the charging force is assumed to equal the spring restoring force. The spring deflection under the system self-weight is neglected in this analysis as jumping robots typically have a higher charging force than their body weight. For example, in \citep{Jumproach_Jung_2019}, the ratio of the peak force to the system weight exceeds 50 and in \citep{NatureJ_Hawkes_2022}, this ratio surpasses 400. 

For reference,  an ideal constant force system produces a constant spring restoring force equal to the peak force and independent of the spring displacement. Its elastic potential energy storage is $PE_{max}=F_{max}d$. \cref{sec:theoretical_models} will examine the four spring-linkage models (\cref{fig:translational_spring_model}) and the energy they store, in comparison to the idealised system.

\section{Theoretical models}
\label{sec:theoretical_models}
The elastic potential energy that can be stored within a spring-linkage system is determined by the locations at which the spring(s) is connected to the linkage, the natural spring length and stiffness, and also the peak spring charging force. \cref{fig:spring_charging_phase}a is a generalised example of a rhomboidal linkage with multiple translational and rotational springs. This section will provide analytical expressions for the required spring charging force in the generalised spring arrangements. It will introduce four spring arrangement models that yield simple analytical expressions for the required spring charging force as a function of the displacement and the spring stiffness. Note that there is no generalised analytical solution to the elastic energy storage (the integral of spring charging force and displacement) of the spring-linkage except for the specific spring arrangements, such as the vertical (\cref{sec:Vertical_spring}), horizontal (\cref{sec:Horizontal spring}) and rotational springs (\cref{sec:Rotational_spring}). \cref{sec:Result_springs} will present the numerical solutions for the elastic energy stored in these four models and study the influence of multiple springs with the same or different spring orientations on energy storage.

\begin{figure}[ht]
\centering
\includegraphics[scale=1]{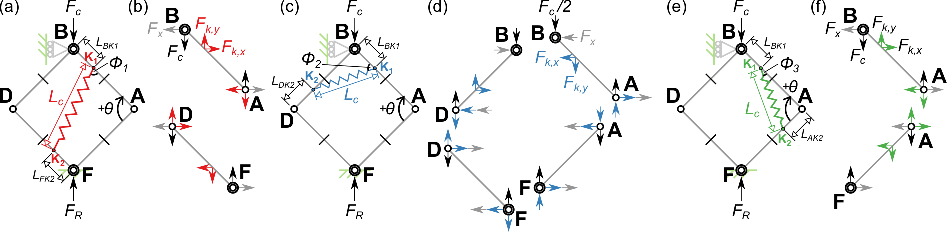}
\caption{The quasi-static models and free-body diagrams of a spring-linkage formed by a rhomboidal linkage and a translational spring attached to links: (a)-(b) BA and FD, (c)-(d) BA and BD, and (e)-(f) BA and FA.  $F_c$ is the spring-charging force, $F_k$ is the spring restoring force and $F_R$ is the ground reaction force.}
\label{fig:translational_spring_model}
\end{figure}

To charge a translational spring that is connected to a rhomboidal linkage the ends of the spring must be connected to two different links. The three possible combinations of the pairs of links that the spring can connect to are:  (1) BA and FD (\cref{fig:translational_spring_model}a), (2) BA and DB (\cref{fig:translational_spring_model}c), and (3) BA and AF (\cref{fig:translational_spring_model}e); note that for each spring arrangement there is an equivalent system where the spring is mirrored about the line BF. 

The \textit{spring positioning ratios}, $\gamma_i$, are length ratios that define the orientation of the spring (\cref{fig:translational_spring_model}): $\gamma_1=L_{BK1}/L_{BA}=L_{FK2}/L_{FD}$, $\gamma_2=L_{BK1}/L_{BA}=L_{DK2}/L_{DB}$ and $\gamma_3=L_{BK1}/L_{BA}=L_{AK2}L_{AF}$, where $L_{BK1}$ is the length of B to K$_1$, $L_{FK2}$ is the length of F to K$_2$, $L_{DK2}$ is the length of D to K$_2$, $L_{AK2}$ is the length of A to K$_2$, $L_{BA}$ is the length of link BA, $L_{FD}$ is the length of link FD, $L_{DB}$ is the length of link DB and $L_{AF}$ is the length of link AF. The angle between the spring and link BA is donated as $\phi_{i}$, which can be expressed as a function of $\gamma_i$ and the spring length, $L_c$: $\phi_1=\sin^{-1} (d\sin\theta/2L_c)$, $\phi_2=\sin^{-1} (L_{DK2}\sin\theta/L_c)$ and $phi_3=\sin^{-1}(L_{AK2}\sin\theta/L_c)$. 

If the spring end points are translated along their respective links while retaining the same spring orientation, the force-length profile and elastic energy remain unchanged. For example, in \cref{fig:translational_spring_model}a if $L_{BK1}$ is reduced 50\% and $L_{FK2}$ is increased 50\% the spring translates, the angle $\phi_1$ is unchanged, and the maximum elastic potential energy that can be stored in the system for a given peak charging force remains unchanged. However, if the orientation of the spring were to change, or if the end points were to be connected to different links, the elastic potential energy storage would change. Therefore the value of $\gamma_i$, and the pair of links being connected, are the two pieces of information required to characterize the system behaviour.

\subsection{Model A: Translational spring connecting links BA and FD}
\label{sec:Model_A}
The first model (\cref{fig:translational_spring_model}a, b) places the translational spring with its ends, K$_1$ and K$_2$, connecting links BA and FD. The spring charging force is expressed as,
\begin{equation}
    F_c=\frac{k\Delta L_c}{\sin\theta} 2\sin\frac{\theta}{2}[(1-\gamma_1)\sin(\frac{\theta}{2}+\phi_1)\cos\frac{\theta}{2}+\gamma_1\cos(\frac{\theta}{2}+\phi_1)\sin\frac{\theta}{2}],\label{eq_Fc_Model_A}
\end{equation}
where $\Delta L_c$ is the spring deflection from its natural length and can be expressed as a function of the knee angle and $\gamma_1$,
\begin{equation}
    \Delta L_c=L_{c,ini}-L_c=\frac{d}{2}[\sqrt{(1-2\gamma_1)(1-2\gamma_1-2\cos\theta_{ini})+1}-\sqrt{(1-2\gamma_1)(1-2\gamma_1-2\cos\theta)+1}].\label{eq_Lc_Model_A}
\end{equation}

The elastic potential energy storage of the spring is given as $EPE=\int F_c\,\text{d}y=\int F_k\,\text{d}L_c$.Note that this model includes two particular spring arrangements that have been used in previous jumping robots: vertical springs ($\gamma_1=0$, $L_{BK1}=L_{FK2}=0$) and horizontal springs ($\gamma_1=1$, $L_{BK1}=L_{FK2}=d/2$), which are discussed in \cref{sec:Vertical_spring} and \cref{sec:Horizontal spring}.

\subsubsection{Vertical spring linkage}
\label{sec:Vertical_spring}
The \textit{vertical spring} linkage is a linear translational spring with its spring axis aligned with the spring charging force, and its ends. An example is a spring with its ends connected to joints H and F of the rhomboidal linkage. The spring-charging force is,
\begin{equation}
    F_c=kd(\sin\frac{\theta_{ini}}{2}-\sin\frac{\theta}{2}).\label{eq_Fc_V}
\end{equation}

The elastic potential energy stored in the spring is $EPE=\int F_c\,\text{d}y=\int F_k\,\text{d}L_c$, so that,
\begin{equation}
    EPE=\frac{F_{max}d}{2}(\sin\frac{\theta_{ini}}{2}-\sin\frac{\theta}{2}).\label{eq_EPE_V}
\end{equation}

A quasi-static model of this system behaves identically to a linear spring system with a prismatic linkage; it has a linear force-length profile (\cref{fig:nF_ny}a) and stores exactly 50\% energy of the idealised spring (\cref{fig:nF_ny}b).    The vertical spring linkage is included here for completeness and for direct comparison with the nonlinear spring linkages presented in this paper.

\subsubsection{Horizontal spring linkage}
\label{sec:Horizontal spring}
The \textit{horizontal spring} linkage is a linear translational spring placed in a rhomboidal linkage with its spring axis is perpendicular to the spring charging force vector. In practice this configuration can be created by connecting an extension spring between the knee joints of the rhomboidal linkage mechanism, and has been demonstrated in various jumping robots \citep{JPL_Hale_2000,PEAnSEA_Hong_2020,JPL_Burdick_2003,FlapJ_Truong_2019,Jumproach_Jung_2016,FleaJ_Koh_2013,ScoutJ_Zhao_2009,ESJ_Bai_2018,Multimobat_Woodward_2014,CLOVER_2022,SurveillanceJ_Song_2009,SphericalJ_Chang_2022}. The spring-charging force is,
\begin{equation}
    F_c=kd \tan\frac{\theta}{2}(\cos\frac{\theta}{2}-\cos\frac{\theta_{ini}}{2}).\label{eq_Fc_H}
\end{equation}

Note that there is a singularity in \cref{eq_Fc_H} at $\theta_{ini}=180\degree$, meaning that the mechanism cannot be compressed from a fully upright standing posture by applying a vertical force; in practice, the compression can be started at an angle slightly lower than $\theta_{ini}=180\degree$.

The knee angle at which the peak charging force is required, $\theta_{ES,P}$, can be solved analytically by differentiating \cref{eq_Fc_H} with respect to the knee angle:
\begin{equation}
    \theta_{ES,P}=2\cos^{-1}[(\cos\frac{\theta_{ini}}{2})^{1/3}].\label{eq_theta_ESP}
\end{equation}

The elastic potential energy stored in the spring is $EPE=\int_{\theta_{end}}^{\theta_{ini}} F_k\,\text{d}\theta$, which is expressed as,
\begin{equation}
  EPE =\begin{cases}
    \frac{F_{max}d(\cos\frac{\theta{end}}{2}-\cos\frac{\theta_{ini}}{2})^2}{2\cos\frac{\theta_{ES,P}}{2}(\cos\frac{\theta_{ES,P}}{2}-\cos\frac{\theta_{ini}}{2})}, & \text{if } \theta_{end} \leq \theta_{ES,P} \\
    \frac{F_{max}d(\cos\frac{\theta{end}}{2}-\cos\frac{\theta_{ini}}{2})}{2\cos\frac{\theta_{end}}{2}}, & \text{if } \theta_{end} \geq \theta_{ES,P}
  \end{cases}
\label{eqn_EPE_H}
\end{equation}

\begin{figure}[t]
\centering
\includegraphics[scale=1]{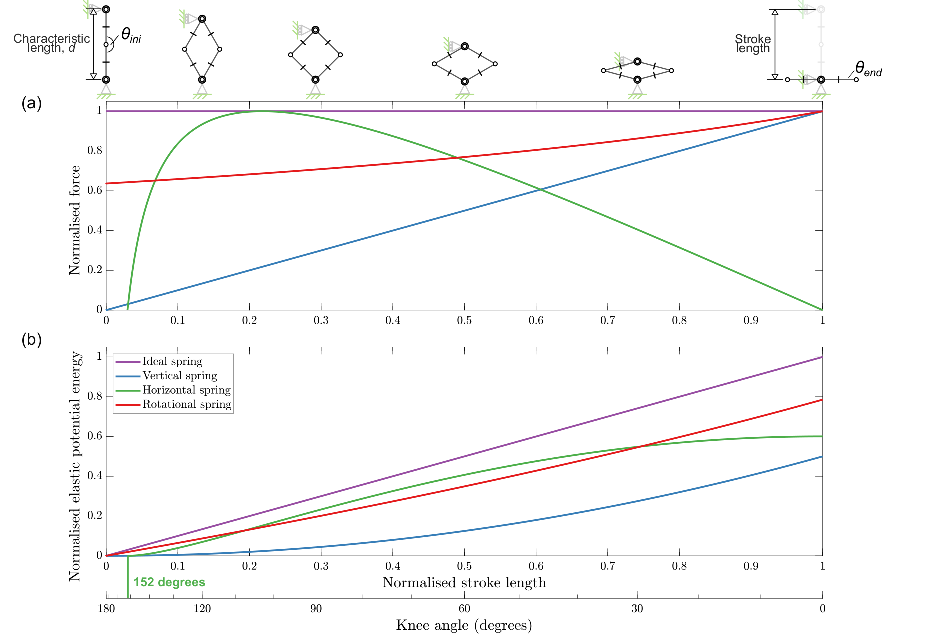}
\caption{The (a) motor force and (c) stored elastic energy of the rhomboidal linkage with the three different spring arrangements throughout the spring-charging process. The initial angles for each model are chosen to maximise the stored elastic potential energy for a given peak motor force and system size. To avoid the singularity in \cref{eq_Fc_V,eq_Fc_Model_D}, both the rotational spring (red) and the vertical spring (blue) systems are initially charged at $\theta_{ini}=179.9\degree$; the horizontal spring (green) is charged at  $\theta_{ini}=152\degree$. The ideal spring system assumes the compression force to remain constant and equal to the peak throughout the entire charging phase, which achieves the maximum amount of elastic energy. Note that the motor force is normalised by the peak motor force and the energy is normalised by the maximum elastic energy stored in the ideal spring system. The compressed stroke length (defined as the displacement of the body from upright, where $\theta_{ini}=180\degree$) is normalised by the characteristic length of the linkage mechanism.}
\label{fig:nF_ny}
\end{figure}

The horizontal spring has to be charged at 152$\degree$ so to obtain the maximum energy. During the spring-charging phase the force increases when the linkage is compressed before reaching a peak at the knee angle of 103$\degree$ (\cref{fig:nF_ny}a, b). As the horizontal spring continues to extend the force reduces, and at maximum compression the spring stores nearly 60\% of the ideal spring energy. The nonlinear force-displacement profile was cited by \citep{JPL_Hale_2000} as a means of increasing power output at the early stages of a jump to negate the effect of the premature take-off in reducing jump height that has been observed for vertical spring robots in \citep{JPL_Fiorini_1999}. However, \citep{LO_Dynamics_2023} will show the early take-off is caused by the centripetal force acting on the rotational body and is unrelated to the spring force profile. From the results shown in \citep{LO_Dynamics_2023}, the improvement in jump height noted in \citep{JPL_Hale_2000} when using horizontal instead of vertical springs is proposed as being due to the increase in stored elastic potential energy, and not due to premature take-off.

\subsection{Model B: Translational spring connecting links BA and DB}
\label{sec:Model_B}
In the second model (\cref{fig:translational_spring_model}c, d), the translational spring ends K$_1$ and K$_2$ connect links BA and DB, respectively. This configuration has yet to be shown in any existing jumping robot design except for the case of the horizontal spring ($\gamma_2=0.5$). The spring charging force is given as,
\begin{equation}
    F_c=\frac{k\Delta L_c}{\sin\theta} \sin\frac{\theta}{2}[(2\gamma_2-1)\sin(\phi_2-\frac{\theta}{2})\cos\frac{\theta}{2}+\gamma_2\cos(\phi_2-\frac{\theta}{2})\sin\frac{\theta}{2}],\label{eq_Fc_Model_B}
\end{equation}
where $\Delta L_c$ is the change of the spring length from its natural length expressed as,
\begin{equation}
    \Delta L_c=L_{c,ini}-L_c=\frac{d}{2}[\sqrt{2\gamma_2(\gamma_2-1)(1-\cos\theta_{ini})+1}-\sqrt{2\gamma_2(\gamma_2-1)(1-\cos\theta)+1}].\label{eq_Lc_Model_B}
\end{equation}
when $\gamma_2=0.5$, the system is representative of the horizontal spring system.

\subsection{Model C: Translational spring connecting links BA and AF}
\label{sec:Model_C}
The third model (\cref{fig:translational_spring_model}e-f) uses a linear translational spring with ends K$_1$ and K$_2$ connecting links BA and AF, respectively. This spring configuration has been used in \citep{FlapJ_Truong_2019}. The spring-charging force is,
\begin{equation}
    F_c=\frac{k\Delta L_c}{\sin\theta} \sin\frac{\theta}{2}[\sin(\phi_3+\frac{\theta}{2})\cos\frac{\theta}{2}+(2\gamma_3-1)\cos(\phi_3+\frac{\theta}{2})\sin\frac{\theta}{2}],\label{eq_Fc_Model_C}
\end{equation}
where $\Delta L_c$ is the change of the spring length from its natural length expressed as,
\begin{equation}
    \Delta L_c=L_{c,ini}-L_c=\frac{d}{2}[\sqrt{2\gamma_3(\gamma_3-1)(1+\cos\theta_{ini})+1}-\sqrt{2\gamma_3(\gamma_3-1)(1+\cos\theta)+1}].\label{eq_Lc_Model_C}
\end{equation}
when $\gamma_3=0.5$, the system is representative of the vertical spring system.

\begin{figure}[t]
\centering
\includegraphics[scale=1]{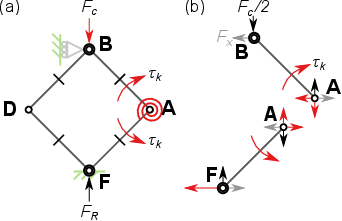}
\caption{The quasi-static models and free-body diagrams of a spring-linkage formed by a rhomboidal linkage and rotational springs at the knee joints A and D.  $F_c$ is the spring-charging force, $F_k$ is the spring restoring force and $F_R$ is the ground reaction force.}
\label{fig:rotational_spring_model}
\end{figure}

\subsection{Model D: Rotational spring}
\label{sec:Rotational_spring}
In the fourth model, a linear rotational spring is placed at joint A (\cref{fig:rotational_spring_model}a), as demonstrated in \citep{TAUB_Zaitsev_2015}. By inspecting \cref{fig:translational_spring_model}b, the charging force is calculated by dividing the rotational spring torque by the moment arm of the applied force (the projection of a link’s length on axis AD) as,
\begin{equation}
    F_c=\frac{2k_r(\theta_{ini}-\theta)}{d\cos\frac{\theta}{2}},\label{eq_Fc_Model_D}
\end{equation}
where $k_r$ is the rotational spring stiffness. As with the horizontal spring, there is a singularity in \cref{eq_Fc_Model_D} when $\theta=180\degree$. The elastic potential energy stored in the spring is given as $EPE=\int F_c\,\text{d}y=\int \tau_k,\text{d}\theta$:
\begin{equation}
    EPE=\frac{1}{4}F_{max}d(\theta_{ini}-\theta_{end})\cos\frac{\theta_{end}}{2}.\label{eq_EPE_Model_D}
\end{equation}

The rhomboidal linkage with rotational springs can store up to 78\% of the elastic potential energy of the idealised system when $\theta_{ini}\rightarrow\pi$ and $\theta_{end}= 0$, as shown in \cref{fig:nF_ny}b. The force-length and energy-length profiles (\cref{eq_Fc_Model_D} and \cref{eq_EPE_Model_D}) of this spring-linkage configuration are independent of the locations and numbers of the rotational springs. For example, placing a pair of rotational springs at joints B and F (as in \citep{MSU_Zhao_2013,Jumproach_Jung_2019}]) or using an elastic cage \citep{NatureJ_Hawkes_2022,Jollbot_Armour_2008} would yield the same force and energy relationship during charging. In practice, this feature allows the system to redistribute the mass towards the top of the system while retaining the same energy storage, which is shown in \citep{LO_Dynamics_2023} as an approach to increase the elastic-kinetic energy efficiency.

\section{Experimental setup of nonlinear spring characterisation}
\label{sec:Exp_general}
A physical experiment was conducted to validate the quasi-static models. The experiment measured the force-displacement relationship of a rhomboidal linkage with rotational springs and horizontal springs. 

\subsection{Experimental model}
\label{sec:Exp_model}
\begin{figure}[b]
\centering
\includegraphics[scale=1]{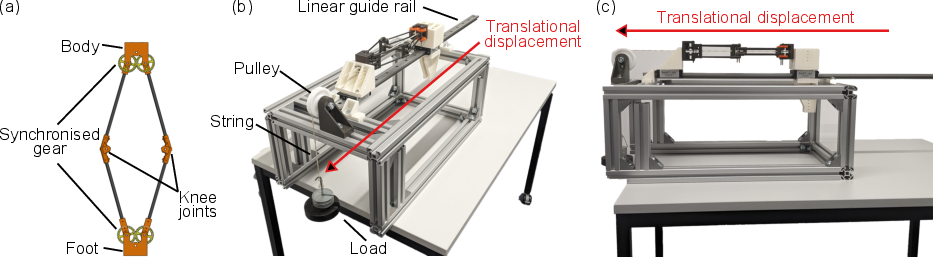}
\caption{(a) CAD model of the experimental model and the setup of the experiment shown in (b) isometric and (c) side views.}
\label{fig:exp_setup}
\end{figure}

The experimental model is a closed-chained linkage mechanism formed by four links, two knee joints, a body and a foot (\cref{fig:exp_setup}a). Each link is a carbon fibre rod measuring 15cm long, with a 6 mm diameter circular cross-section. The ends of the rods are connected to a knee joint and either a body or foot component. The knee joint, fabricated from polylactic acid (PLA), also serves as the spring housing for the translational spring and the rotational springs. The body and the foot components, also made of PLA, function as mounting platforms for attaching the linkage to the test rig (\cref{fig:exp_setup}b), which will be introduced later.  The design has a single degree of freedom and can only articulate translationally following the 1m long Igus T linear guide rail (\cref{fig:exp_setup}c). Inspired by \citep{JPL_Hale_2000,Jumproach_Jung_2019}, this is achieved by the pairs of synchronised gears located at the end of each link. Low friction bearings MF52ZZ are installed in each revolute joint with planet-based lubricants to minimise the frictional effect during the articulation. 

\subsection{Spring arrangements}
\label{sec:Exp_Spring_arrangements}
Two experiments were conducted with different springs. The first experiment used a linear translational spring in the \textit{horizontal spring} linkage arrangement. The spring has a natural length of 8cm and stiffness of 200Nm$^{-1}$ \citep{Extension_spring}. The spring is attached to the knee joints of the linkage. The natural spring length gives the system an initial standing angle, $\theta_{ini}= 164\degree$; there was no observable deflection caused by the system weight. The maximum stretched length of the spring gives the terminal angle, $\theta_{end}= 44\degree$.

The second experiment used a pair of rotational springs attached to the knee joints. Each spring has a spring stiffness of 0.7Nmrad$^{-1}$ and a natural angle of 178$\degree$ \citep{Torsion_spring}. The initial angle of the second experiment is equal to the natural angle of the spring; again, there was no observable deflection under the system weight. The terminal angle of the system, is 28$\degree$, where it cannot be compressed further due to contact being made between links BA and FA.  

\subsection{Experiment setup}
\label{sec:Exp_setup}
The linkage was mounted on a 1m long Igus T linear guide rail, with the translational degree of freedom perpendicular to gravity, measured using a sprit level. A supporting structure was made using extruded aluminium profile beams (\cref{fig:exp_setup}b). The rig was placed on a flat surface of a laboratory bench. The entire structure was assumed to be perfectly rigid. A string is connected to the body of the experimental model to a load via a pulley; the string and load path are assumed parallel to the local gravity vector. The spring-linkage is compressed by the load. The load was incremented in 100gram steps, and the displacement of the system was recorded visually (to the nearest 1mm) using a ruler aligned parallel to the guide rail. Each experiment was repeated 10 times.

Error bars of measured displacements were calculated by dividing the standard deviation by the square root of the mean of the measurements using the inbuilt function, \textit{errorbar()}, in Matlab. During the experiments there were no observable deformations of the structure of the linkage or the rig, and so measurements of the system deflections under load were assumed to be due to deflections of the spring alone.

\section{Numerical models}
\label{sec:Num_models}
The normalised elastic energy stored in the analytical models in \cref{sec:theoretical_models} will be presented in \cref{sec:Result_springs} to show the influence of the spring orientation and number of springs on the energy storage. The elastic energy was computed by the numerical integration of the spring charging force (\cref{eq_Fc_Model_A}, \cref{eq_Fc_Model_B} and \cref{eq_Fc_Model_C}) over the displacement of the body (\cref{eq_y}) using the Matlab inbuilt function, \textit{cumtrapz()}, with 1000 data points space uniformly between $\theta_{ini}$ and $\theta_{end}$. Further increase in data points resulted in less than 0.1\% change in all results. 

A second numerical multibody dynamics model was developed using the Matlab Simulink (Simscape Multibody toolbox library ) to verify the analytical models in \cref{sec:theoretical_models}. This toolbox simulates the system dynamics under inertial-, applied external-, and spring-forces and moments. By default this toolbox simulates system dynamics, including inertial effects. Here, a quasi-static solution was obtained by incorporating a PD controller to apply a charging force based on a setpoint of required system displacement. The derivate term ensured that the system velocity and acceleration tended to zero for the given setpoint, and hence inertial effects could be neglected, making the model equivalent to the analytical models. The geometry of the system was defined as being equal to that of the experiment. The link masses are each set as 0.00001kg, so as to be negligible for consistency with the analytical model. The modelling parameters for translational spring are $k=200$Nm$^{-1}$, $\theta_{ini}=164\degree$ and $\theta_{end}=0.01\degree$, and for the rotational spring, $k_r=0.7$Nmrad$^{-1}$, $\theta_{ini}=170\degree$ and $\theta_{end}=0.01\degree$. Simulations are computed using a fixed timestep step solver (ODE3) with a timestep of 0.0001s; further reduction in timestep yielded less than 0.1\% change in all results. 

\section{Results}
\label{sec:Results}
This section will show the influence of the spring numbers and locations on the energy storage capacity of the spring-linkage system. It will firstly present the experiments and numerical models to validate the analytical models of the rhomboidal linkage with the horizontal spring or rotational springs.  The remainder of the results will show that the addition of a single translational spring creates a nonlinear spring that can store up to 65\% energy of an idealised constant force spring system. This is 13\% lower than that achieved with rotational springs  (see \cref{sec:theoretical_models}). Two example ideal constant force springs will be illustrated by combining the rhomboidal linkage and multiple springs. 

\begin{figure}[t]
\centering
\includegraphics[scale=1]{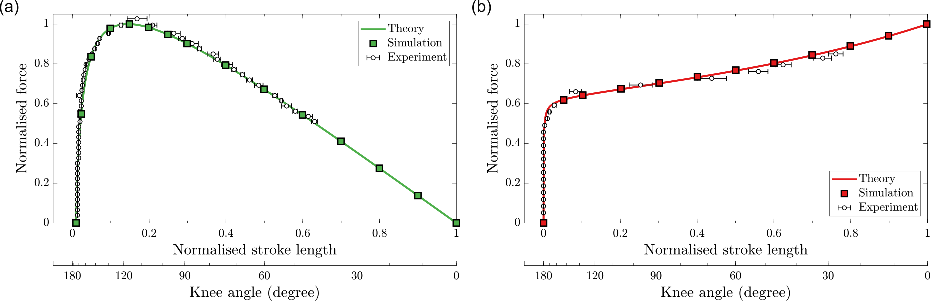}
\caption{Validation of the (a) horizontal and (b) rotational springs theories with the Simscape simulation model and the static experiment. The input force is normalised by the peak force in the theoretical model.}
\label{fig:exp}
\end{figure}

\subsection{Experimental validation of the theoretical models}
\label{sec:Result_valid}
The experimental data in \cref{fig:exp} follow the trends predicted by the numerical simulation in \cref{sec:Num_models} and the analytical models in \cref{sec:Vertical_spring} and \cref{sec:Horizontal spring}. The numerical and analytical models are in agreement to within the machine precision of the computation. In the case of the horizontal spring (\cref{fig:exp}a) the experimentally measured peak spring charging force is 2.6\% higher than the theoretical prediction. The minimum knee angle in the experiment is $\theta=44\degree$, limited by the maximum stretch length of the spring. 

For the rotational spring (\cref{fig:exp}b) the measured force is 2.2\% lower than the predicted value at $\theta=28\degree$, which is the minimum knee angle limited by the physical constraints of the linkage. Potential sources of discrepancy between the theoretical and experimental results include nonlinearity in the spring torque-displacement and force-displacement behaviour. However, the error magnitudes are regarded as being sufficiently small to assume that the analytical and numerical models are representative of the quasi-static behaviour of the linkage systems that they model.

\subsection{The effect of spring orientation and number of springs on energy storage}
\label{sec:Result_springs}

\begin{figure}[t]
\centering
\includegraphics[scale=1]{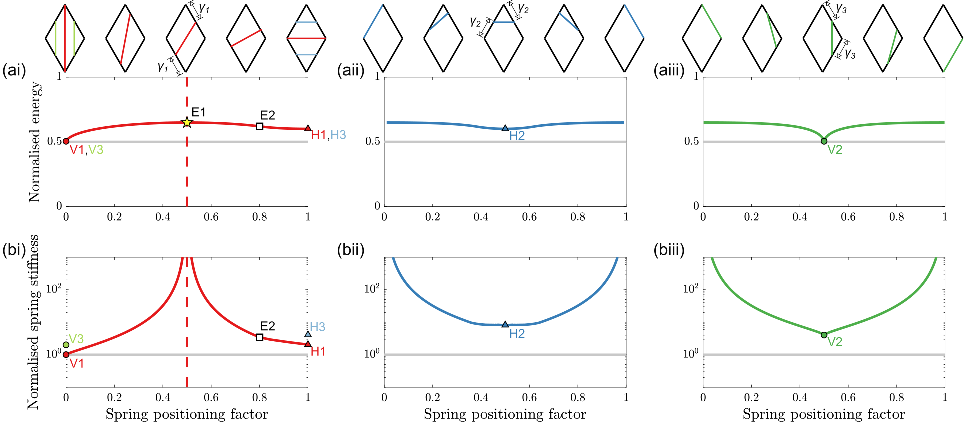}
\caption{The (ai-aiii) energy and (bi-biii) required spring stiffness of the rhomboidal linkage with translational spring placed at the orientation defined by the factors, $\gamma_i$. The energy capacity of the spring-linkage system depends on the spring orientation for a given actuation force and system length, and is independent of the number of springs at the same orientation, which only affects the required spring stiffness of the system. These can be seen from the examples of vertical springs (V1-V3) and horizontal springs (H1-H3). Note that the energy and the spring stiffness are normalised by the peak actuation force and the characteristic length of the system. For reference, the light grey lines in (a) and (b) represent the normalised energy and stiffness of a linear translational spring.}
\label{fig:spring_orientation}
\end{figure}

\cref{fig:spring_orientation} shows the elastic potential energy storage and the required spring stiffness of a rhomboidal linkage system with different linear spring placements. The maximum spring charging force and system characteristic length is constant for all cases. The spring stiffness is selected to enable full compression of the system by the given charging force.  
The minimum elastic potential energy is stored when the spring is placed vertically (V1 in \cref{fig:spring_orientation}). This spring orientation incurs greater spring displacement than in other orientations. As mentioned in \cref{sec:Vertical_spring}, the linkage mechanism is redundant in this case because the charging force and the spring axis are collinear. There is no horizontal force or torque applied by the linkage so the spring is unaffected by the presence of the linkage. The vertical spring has the lowest stiffness of all the spring orientations tested to be fully compressed by a given charging force. 

The maximum elastic potential energy storage of the rhomboidal linkage is realised as the spring orientation tends to be parallel with any link in the mechanism (e.g. E1 in \cref{fig:spring_orientation}ai). The energy stored is up to 65\% of the energy stored in an idealised spring. However, as the spring tends to this orientation its displacement tends to zero throughout the charging phase and the required spring stiffness tends to infinity. So in practical terms this optimum energy condition cannot be realized. A normalised energy of 62\% of the idealised system can be achieved with a position factor of 0.8 and normalised stiffness of 3.5, as shown in E2 in \cref{fig:spring_orientation}ai and bi. 

It should be noted that the spring stiffness would also influence its mass, which is not taken into consideration here. It is plausible that using a stiffer spring would increase the system mass, and potentially require additional mass for spring housings \citep{FlapJ_Truong_2019}, which in turn would reduce the elastic-kinetic energy conversion efficiency \citep{LO_Dynamics_2023}. In comparison, a horizontal spring (around 60\% of ideal spring energy) charges 5\% less energy than the maximum, but it can be easily installed to the knee joints of the linkage (e.g. an elastic band connected to the revolute joints \citep{Jumproach_Jung_2016,CLOVER_2022}) and requires a less stiff spring (e.g. H1 in \cref{fig:spring_orientation}ai and bi). A practical problem with this approach is the lack of system compactness and protection if the spring stretches across the middle of a robotic mechanism.

Changing the spring in position, or the links to which it connects, while retaining its orientation does not change the elastic potential energy storage of the system. This is because the required spring stiffness will vary to compensate for the change in the available spring displacement of each spring placement. The caveat in this argument for real-world springs (e.g. helical spring) is that they have finite minimum lengths, which limits the extent to which spring length and stiffness can be traded off.

Increasing the number of springs of a given orientation does not change the energy storage capacity of a system. This is shown in V3 and H3 in \cref{fig:spring_orientation}ai and bi, where the spring number is doubled but each spring is also 50\% less stiff compared to V2 and H2; the normalised energy remains unchanged under a given force and characteristic length. This finding is interesting given previous research into jumping robots that use multiple springs (e.g. \citep{NatureJ_Hawkes_2022,Jumproach_Jung_2016,Jumproach_Jung_2019,MSU_Zhao_2013,LO_2021}) to reach a desired stiffness. It is hypothesised that the intention of these works was to create a bespoke spring stiffness so as to enable full compression of the system by their given motor force.

\begin{figure}[ht]
\centering
\includegraphics[scale=1]{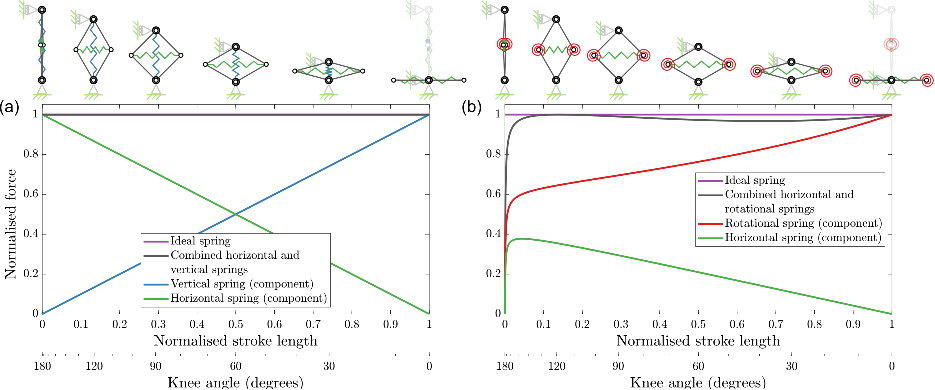}
\caption{The idealised systems generated by the rhomboidal linkages with the combination of different spring types and allocations: (a) vertical spring and horizontal spring, and (b) rotational spring and horizontal spring. The second approach has already been implemented in \citep{NatureJ_Hawkes_2022,Jumproach_Jung_2019}}
\label{fig:combined_spring}
\end{figure}

The key to changing the energy storage capacity of a spring-linkage is to use multiple springs with different spring orientations, or to use different spring types. By doing so, the resultant force-length profile of the multi-spring-linkage mechanism is the superposition of force-length profiles of the individual spring-linkages, as shown in \cref{fig:combined_spring}. Therefore, it is conceptually possible to create an idealised constant force spring using a rhomboidal linkage with a vertical spring and horizontal spring with the same stiffness (\cref{fig:combined_spring}a). The gradients of the force-stroke length lines of the two springs are of equal magnitude but opposite sign. The summation of the two yields a constant force system, equivalent to the ideal springs referred to throughout this paper (though noting the singularity of the horizontal spring at $\theta=180\degree$ is still present in the combined system). Implementing this system in a practical robot would likely utilise a helical compression spring for the vertical spring, which would require a means of preventing buckling.

\cref{fig:combined_spring}b illustrates a multi-spring system that is straightforward to implement in a practical jumping robot, and stores 97\% of the elastic energy of the idealised system. The system uses rotational springs at the knees, and horizontal spring. This approach has been implemented in a previous jumping robot \citep{Jumproach_Jung_2019}, however, it was recorded to only store 65\% of the ideal system energy (Fig. 11 in \citep{Jumproach_Jung_2019}). The reason for this is that the stiffness of the chosen springs did not yield a constant force-displacement profile.  In practice, the stiffness and the number of springs should be selected to match with the maximum force of the selected spring-charging motor. A near constant force-displacement profile was achieved in \cite{NatureJ_Hawkes_2022}, which combined a cage-like structure with multiple horizontal springs. The resultant system jumped 33m high and set the jump height record of jumping robots. It is noteworthy that this robot only utilises 66\% of its characteristic length for spring charging (as shown in \cref{fig:intro}fvi). 

A key finding is that the present analysis can utilize a simple mechanics model to capture the overall energetics trends seen in previous jumping robots, which have a range of intricate structures (e.g. \citep{NatureJ_Hawkes_2022,Jollbot_Armour_2008,Springtail_Ma_2021}) The models also provide a means of proposing how the mechanical-elastic energy conversion efficiency can be improved in previous nonlinear spring-driven jumping robots.. This paper shows that it is mechanically relatively straightforward to develop spring-linkage systems with energy storage capacities that surpass that of a simple linear spring . This is particularly significant given recent studies that have predicted sizing limits on jumping actuator types \citep{Ilton_2018}, but based their predictions on linear spring systems.

\subsection{Application of the present theory to jumping robot technologies}
\label{sec:Result_application}
The previous sections explored the influence of the spring numbers and orientations on the elastic energy storage of spring-linkage models. This section aims to bridge the theoretical discussion with practical implications by showcasing the real-world effect of the theories developed in this paper.  Hence, this section will compare the energy storage between the previous spring-driven jumping robots \citep{CLOVER_2022,FleaJ_Koh_2013,Jumproach_Jung_2016,Jumproach_Jung_2019,MSU_Zhao_2013,NatureJ_Hawkes_2022} and the theoretical spring-linkage models. It will also predict the potential increase in the jump height of these previous robots through the application of the ideal spring linkage discussed in \cref{sec:Result_springs}.

\begin{figure}[t]
\centering
\includegraphics[scale=1]{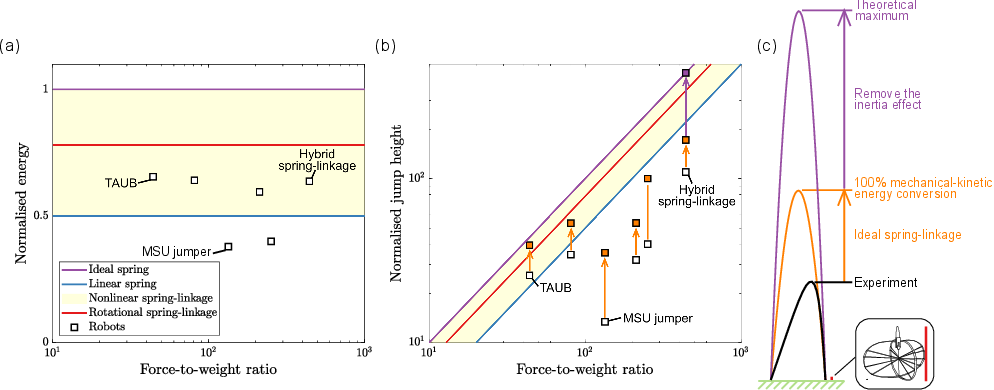}
\caption{(a) The elastic energy storage and (b) the jump height of the previous spring-driven jumping robots with spring-linkage mechanisms \citep{CLOVER_2022,FleaJ_Koh_2013,Jumproach_Jung_2016,Jumproach_Jung_2019,MSU_Zhao_2013,NatureJ_Hawkes_2022}, compared to the spring-linkages, the ideal and linear spring models, , assuming no inertia effect. None of the robots have stored more elastic energy or achieved a higher jump height than the equivalent ideal spring system at a given force-to-weight ratio. (c) The experimental jumping trajectory of the hybrid spring linkage \citep{NatureJ_Hawkes_2022} and the predicted jumping trajectory of the same robot using the ideal spring-linkage and without the inertia and aerodynamic effects. Note that the elastic energy is normalised by the elastic energy stored in the ideal spring and the jump height is normalised by the characteristic length of the system. The force-to-weight ratio is the ratio of the peak spring charging force (e.g. from a motor), $F_{max}$, to the weight of the system. The jump height of the robots is derived from the experimental measurement of the take-off velocity, $v_{to}$, using the formula, $h=\frac{1}{2g}v_{to}^{2}$; the jump height of the theoretical models is derived by its stored elastic energy by the formula, $h=\frac{EPE}{mg}$, with the dynamic effect being neglected. The white makers indicate the original energy storage and jump height of the robots; the orange makers indicate the improved jump height of the robot using the ideal spring-linkage mechanism in this work; the purple markers indicate the theoretical jump height maximum of the robot using the ideal spring-linkage mechanism and without the dynamic effect.}
\label{fig:application}
\end{figure}

\cref{fig:application} shows the elastic energy storage and the jump height of the previous nonlinear spring-driven jumping robots. The results of the ideal and linear springs and the general nonlinear spring-linkage are shown as a reference under the assumption of no inertia effect. The force-to-weight ratio is defined as the ratio of the spring charging force to the weight of a system. The increase in force-to-weight ratio of a jumping robot can be theoretically framed as using a motor with a greater peak force to charge more elastic energy while neglecting the mass increase of the spring or motor. The theoretical jump heights of the theoretical models depicted in \cref{fig:application}b and \cref{fig:application}c  are derived from the elastic potential energy storage, assuming the conservation of energy.  The dynamic effects of the physical system are neglected here. 

As a general overview, the jump height and the elastic energy storage of the robots are lower than the theoretical boundary set by the ideal spring, which has the greatest elastic energy storage for a given spring charging force (\cref{sec:theoretical_models}). Among the robots, the hybrid spring-linkage, which currently holds the record for jump height, stores up to 64\% elastic energy of the ideal spring and jumps up to 110 times its characteristic length \citep{NatureJ_Hawkes_2022}. However, this experimental jump height achievement is only equivalent to 25\% of the theoretical maximum and is even lower than the linear spring at its respective force-to-weight ratio. The losses are attributed to the mechanical-elastic energy efficiency in the spring-charging phase and the elastic-kinetic energy efficiency in the acceleration phase. 

The data in \cref{fig:application}a suggests that the mechanical-elastic energy conversion efficiency in a practical spring-linkage system is independent of the spring-charging force, system weight and characteristic length, in line with the theoretical models presented in \cref{sec:theoretical_models}. However, it is interesting to note that the energy efficiency of the spring-linkage robots included are much lower than their respective theoretical maximum. A notable example is the MSU jumper \citep{MSU_Zhao_2013}. As a rotational spring-linkage, it can store up to 78\% of ideal spring energy but \citep{MSU_Zhao_2013} could only manage 37\%, which is even lower than a linear spring. The main reason of this is the physical constraint imposed by the mechanism design of the robot. The cam design restricts the articulation of the robot and consequently hampers the spring from reaching its full charge. This issue is addressed in another rotational spring-linkage, the TAUB jumper \citep{TAUB_Zaitsev_2015}, as shown in \cref{fig:intro}fiv. But despite that it still reaches only 65\% efficiency, which remains 13\% below the theoretical maximum for a rotational spring linkage. The reason of this is the stiffness of the chosen spring does not match with the peak spring charging force.  Therefore, to effectively implement the design guidelines presented in this paper, a case-specific study would be required to design the articulated mechanism and manufacture bespoke springs tailored to the specific requirements of individual spring-linkage robots. 

In the pursuit of 100\% mechanical-elastic energy conversion efficiency of the existing spring-linkage designs, this work presented a potential route by combining a rhomboidal linkage with the vertical and horizontal springs to form an ideal spring (\cref{sec:Result_springs}). \cref{fig:application}b approximates the effect of implementing this approach on the jump height of the previous jumping robots (shown as orange markers). The results show a substantial increase in the jump height from 50\% to over 160\%. Nonetheless, it is noted that none of the examples reached the theoretical maximum set by the ideal spring-driven system. More interestingly, the improved jump height of some robots remains lower than the linear spring-driven system. For example, the hybrid spring-linkage using the ideal spring-linkage (\cref{fig:application}c) would increase its normalized jump height from 110 to 172, which is around 39\% of its theoretical maximum and is below the linear spring of its respective force-to-weight ratio. The remaining 61\% of the energy loss is attributed to the inefficient elastic-kinetic energy conversion during the acceleration phase, caused by the rotational and unsprung masses in the physical linkage \citep{LO_Dynamics_2023}. While this might appear to be a drawback inherent to the mechanism, it does not imply that the spring-linkage is impractical. Traditional prismatic mechanisms (e.g. \citep{JPL_Fiorini_1999}) also experience energy loss due to unsprung foot mass \citep{LO_Dynamics_2023}. Therefore, while a spring-linkage suffers from inefficient elastic-kinetic energy conversion due to the inertia effect, it can still yield a greater jump height than a physical prismatic linear spring system due to the increased elastic energy storage as shown in \citep{JPL_Hale_2000}. 

\section{Conclusion}
\label{sec:Conclusion}
This paper presented an energetics analysis of nonlinear springs formed by combining rhomboidal linkage and linear springs. The knowledge can be applied to the design of spring-driven jumping robots, as one of the common goals of this class of jumper is to maximise the stored elastic energy in a spring, in order to increase mechanical-kinetic energy conversion efficiency. Jumping performance (e.g. jump height, power) is not dependent on elastic energy storage alone, but is also determined by the conversion of the elastic energy to kinetic energy. However, to make this problem tractable the present paper tackled the first stage of the energy conversion from mechanical work to elastic energy, and \citep{LO_Dynamics_2023} deals with the second stage of converting elastic to kinetic energy.

A rhomboidal linkage with a translational spring orientated nearly parallel with any one of the links stores around 65\% of the energy of an ideal spring. This provides higher elastic energy storage compared to other translational spring arrangements. Also, aligning a spring nearly parallel to a  link offers a compact structural package that protects the spring from being exposed to the dangers of the environment. A caveat with this finding is that as the spring orientation tends to being parallel with a link, the spring displacement reduces and the required spring stiffness increases. This may in turn reduce the overall elastic-kinetic conversion efficiency after inertial effects are considered.

In terms of overall compactness, the rotational spring also stands out as a viable design option, and is shown here to be capable of storing around 78\% of the elastic energy of the ideal spring. The analysis also showed that rotational springs can be applied at any joint, offering the ability to adjust the mass distribution of a jumping system, which is known to be a critical factor in elastic-kinetic energy conversion \citep{LO_Dynamics_2023}.

Combining different classes or springs within a rhomboidal linkage has been shown here to be a means of achieving higher energy storage than with a single spring. This paper formulates two approaches to creating a near-ideal nonlinear spring. The first approach uses a rhomboidal linkage with vertical and horizontal springs to create a constant force spring, with 100\% mechanical-elastic energy conversion. The second approach uses a rhomboidal linkage with rotational and horizontal springs, which stores up to around 97\% energy of an ideal spring. In practice, the challenge of realising these design concepts is sourcing springs with the correct stiffness properties, which would require bespoke spring design as in \citep{NatureJ_Hawkes_2022,Jumproach_Jung_2019}. Also, the exposure of the horizontal spring to the environment may make the design impractical for some jumping robot applications, such as planetary exploration.

\bibliographystyle{elsarticle-num} 
\bibliography{Reference}





\end{document}